%% file: main.tex
\documentclass[10pt,twocolumn,letterpaper]{article}

\usepackage[pagenumbers]{iccv}              


\definecolor{iccvblue}{rgb}{0.21,0.49,0.74}
\usepackage[pagebackref,breaklinks,colorlinks,allcolors=iccvblue]{hyperref}
\usepackage{multirow}
\usepackage{pifont}
\usepackage[normalem]{ulem}
\useunder{\uline}{\ul}{}

\newcommand{\cmark}{\ding{51}}%
\newcommand{\xmark}{\ding{55}}%
\definecolor{deepgreen}{HTML}{009900}

\def\paperID{9135} 
\def\confName{ICCV}
\def\confYear{2025}

\newcommand{\model}{\textsc{Vamba}\xspace}

\title{\model: Understanding Hour-Long Videos with Hybrid Mamba-Transformers}

\author{
Weiming Ren$^{1,4}$, Wentao Ma$^2$, Huan Yang$^{3}$, Cong Wei$^{1,4}$, Ge Zhang$^{1,5}$, Wenhu Chen$^{1,4}$\\
$^{1}$University of Waterloo, $^2$University of Toronto, $^3$Kuaishou Technology, $^4$Vector Institute, $^5$M-A-P\\
{\tt\small\{w2ren,wenhuchen\}@uwaterloo.ca, hyang@fastmail.com}\\
{\normalsize \url{https://tiger-ai-lab.github.io/Vamba/}}
\vspace{-1.5em}
}

\begin{document}
\maketitle
\input{sec/0_abstract}    
\input{sec/1_intro}
\input{sec/3_prelim}
\input{sec/3_method}
\input{sec/4_experiment}

\input{sec/2_related_work}
\input{sec/5_conclusion}

{
    \small
    \bibliographystyle{ieeenat_fullname}
    \bibliography{main}
}

\input{sec/X_suppl}

\end{document}

%% file: sec/0_abstract.tex
\begin{abstract}
State-of-the-art transformer-based large multimodal models (LMMs) struggle to handle hour-long video inputs due to the quadratic complexity of the causal self-attention operations, leading to high computational costs during training and inference. Existing token compression-based methods reduce the number of video tokens but often incur information loss and remain inefficient for extremely long sequences. In this paper, we explore an orthogonal direction to build a hybrid Mamba-Transformer model (VAMBA) that employs Mamba-2 blocks to encode video tokens with linear complexity. Without any token reduction, VAMBA can encode more than 1024 frames (640$\times$360) on a single GPU, while transformer-based models can only encode 256 frames. On long video input, VAMBA achieves at least 50\% reduction in GPU memory usage during training and inference, and nearly doubles the speed per training step compared to transformer-based LMMs. Our experimental results demonstrate that VAMBA improves accuracy by 4.3\% on the challenging hour-long video understanding benchmark LVBench over prior efficient video LMMs, and maintains strong performance on a broad spectrum of long and short video understanding tasks.

\end{abstract}

%% file: sec/1_intro.tex
\section{Introduction}
\label{sec:intro}

The field of large multimodal models (LMMs) has seen tremendous progress in recent years. The seminal work of LLaVA \cite{liu2024visual} successfully transferred the power of autoregressive next-token prediction from large language models (LLMs) \cite{OpenAI_ChatGPT_Website, achiam2023gpt, touvron2023llama, bai2023qwen, jiang2023mistral} to the multimodal domain, establishing a new standard for visual understanding by autoregressively modelling visual and language tokens through causal transformer layers. Since then, transformer-based LMMs have been widely studied and enhanced to tackle a diverse range of tasks, such as high-resolution image understanding \cite{llavanext2024, laurenccon2024matters, wang2024qwen2} and interleaved image-text reasoning \cite{jiang2024mantis, li2024llava, li2024llavaov}. Nevertheless, the quadratic complexity inherent in causal self-attention operations presents significant challenges for long-context inputs, making it difficult for transformer-based LMMs to effectively handle the task of understanding extremely long videos.

\begin{figure}[t!]
    \centering
    
    \begin{subfigure}[b]{0.9\linewidth}
        \centering
        \vspace{-2pt}
        \includegraphics[width=\linewidth]{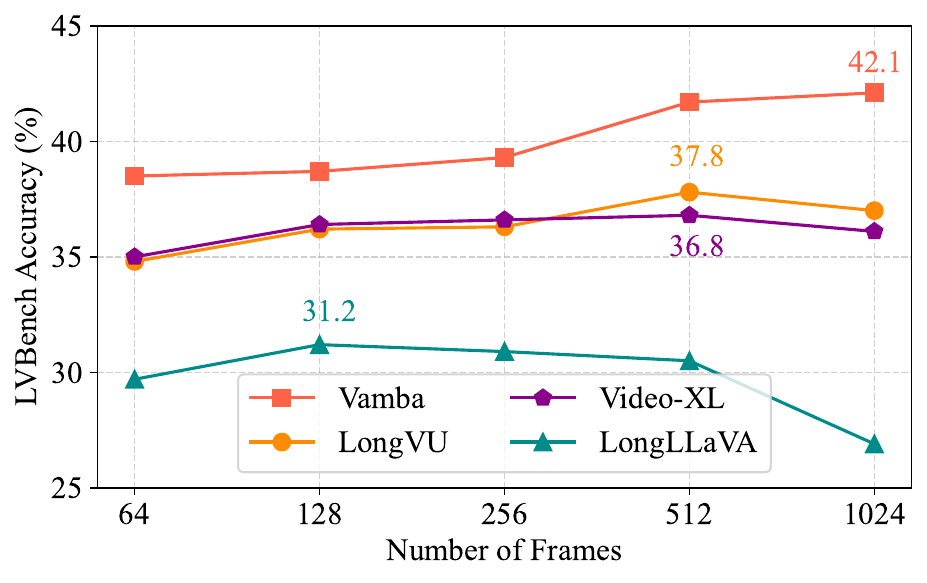}
        \label{fig:lvbench}
    \end{subfigure}
    
    \vspace{-1.5em}
    
    \begin{subfigure}[b]{1\linewidth}
        \centering
        \vspace{-2pt}
        \includegraphics[width=\linewidth]{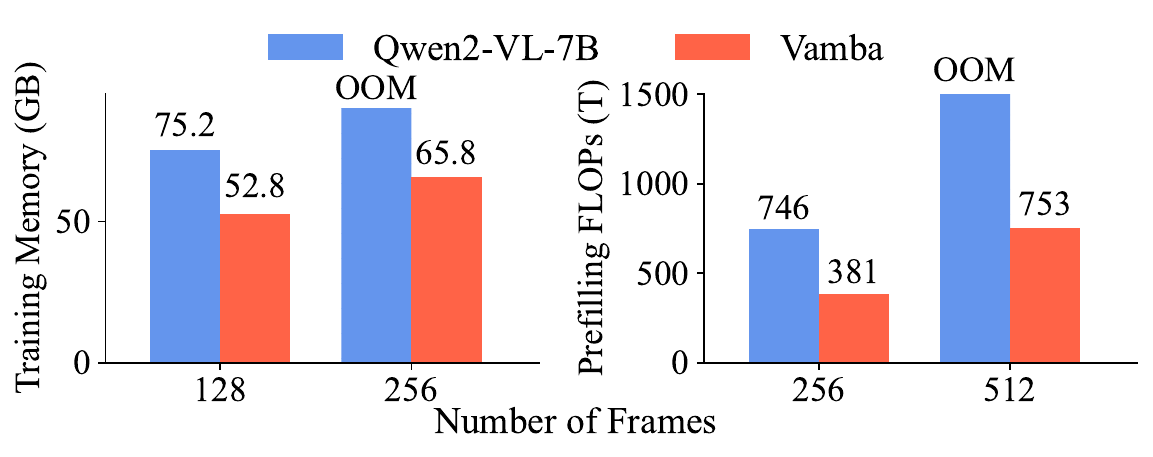}
        \label{fig:compare}
    \end{subfigure}
    \vspace{-3.5em}
    \caption{\model achieves strong long video understanding performance (42.1\% on LVBench \cite{wang2024lvbench}) while being more computationally efficient compared to transformer-based LMMs.}
    \label{fig:teaser}
    \vspace{-1.5em}
\end{figure}

A key challenge for advanced LMMs when processing long video inputs is that they tend to encode each frame into a large number of vision tokens, which leads to substantial computational and memory costs during both training and inference. For example, Qwen2-VL~\citep{wang2024qwen2} can only process 256 frames (360p) on a single GPU, which is far from sufficient for hour-long video understanding. To reduce the computational/memory cost, previous efforts have primarily focused on reducing the vision tokens in the input sequence. Several approaches \cite{zhang-etal-2023-video, li2023videochat, li2024mvbench, fei2024video} utilize a Q-Former \cite{li2023blip} to compress vision tokens. More recent methods, such as LongVU \cite{shen2024longvu} and Video-XL \cite{shu2024video}, partition vision token sequences into chunks and employ adaptive token compression mechanisms to decrease the token count. Another line of work, such as InternVideo2.5~\citep{internvideo2_5} and Orxy-1.5~\citep{liu2024oryx}, employs various mechanisms to evaluate the vision tokens' importance, and thus dropping or merging those less significant tokens \cite{tokenmerging} during training and inference. Despite these advancements, key challenges in LMMs for long video understanding still persist. First, aggressive token reduction can lead to critical information loss, particularly when processing extremely long videos. Second, these methods still suffer from quadratic computational complexity as the number of input frames increases. Third, token reduction-based methods require additional overhead, which might increase the actual wall-clock time.

\input{tables/complexity}

In this study, we investigate an orthogonal direction to previous approaches: instead of compressing the video tokens, we seek to develop an alternative model architecture that improves the efficiency of processing video tokens during training and pre-filling stage of inference. We propose \model, a hybrid Mamba-Transformer model for efficient hour-long video understanding. The key insight of our method is that we can design efficient modules to approximate the causal self-attention operation for both text and video tokens in transformer-based LMMs. In particular, we propose to (1) utilize cross-attentions to update text tokens based on video tokens, which is affordable due to the short length of text tokens, and (2) adopt Mamba-2~\citep{dao2024transformers} to process the massive video tokens with linear complexity.

Assuming a combined input sequence of $M+N$ tokens, where $M$ is the number of video tokens and $N$ is the number of text tokens, we find that $M$ could be at least 100 times larger than $N$ on many long video tasks ($M \gg N$). Our model can reduce the training/pre-filling computational complexity from $O(d(M+N)^2)$ to $O(dMN+d^2M)$, where $d$ is the hidden dimension, according to Table~\ref{tab:complexity}. In practice, this theoretical improvement may not be fully realized due to the hardware under-optimization for Mamba~\citep{gu2023mamba}. Nevertheless, we still observe a more than 50\% reduction in GPU memory usage and FLOPs/runtime during training and inference for long video inputs, as shown in Figure~\ref{fig:teaser}. \model can be efficiently trained using 8$\times$A800 GPUs, whereas other efficient video LMMs such as LongVU \cite{shen2024longvu} and LongLLaVA \cite{wang2024longllava} require 64 and 24 GPUs for training, respectively. By performing a two-stage training, our \model achieves a 4.3\% improvement over the best efficient video LMMs on the challenging hour-scale video understanding benchmark LVBench \cite{wang2024lvbench}. On other long video understanding datasets like Video-MME~\citep{fu2024video}, MLVU~\citep{zhou2024mlvu} and LongVideoBench~\citep{wu2025longvideobench}, \model also achieves top-notch performance.

\noindent Our contributions can be summarized as follows:
\begin{itemize}
    \item [1.] We propose \model, a hybrid Mamba-Transformer model for hour-long video understanding. \model's design features efficient modules such as Mamba-2 blocks and cross-attention layers, effectively reducing the computational overhead of transformer-based LMMs.
    \item [2.] We conduct a comprehensive ablation study and show that employing Mamba-2 blocks and initializing cross-attention weights from pretrained self-attention layers are crucial for achieving high performance in \model.
    \item [3.] Our extensive evaluations demonstrate that \model achieves strong video understanding capabilities. Specifically, \model achieves 4.3\% improvement over state-of-the-art efficient video LMMs on the challenging hour-long benchmark LVBench.
\end{itemize}

%% file: tables/complexity.tex
\begin{table}[]
\centering
\small
\caption{Theoretical time and memory complexity for transformer-based LMMs and \model in pre-filling. $M$ and $N$ represent the number of vision and text tokens. $d$ denotes the hidden dimension.}
\setlength\tabcolsep{7 pt}
\begin{tabular}{l|c|c}
\toprule
Model                  & Time Complexity & Memory Complexity \\ \midrule
Transformer            &     $O(d(M+N)^2)$             &   $O((M+N)^2)$             \\
\model                 &   $O(dMN+d^2 M)$              &   $O(MN)$                \\
\bottomrule
\end{tabular}
\label{tab:complexity}
\vspace{-1.5em}
\end{table}

%% file: sec/3_prelim.tex
\section{Preliminaries}

\begin{figure*}[!t]
  \centering
  \includegraphics[width=0.9\textwidth]{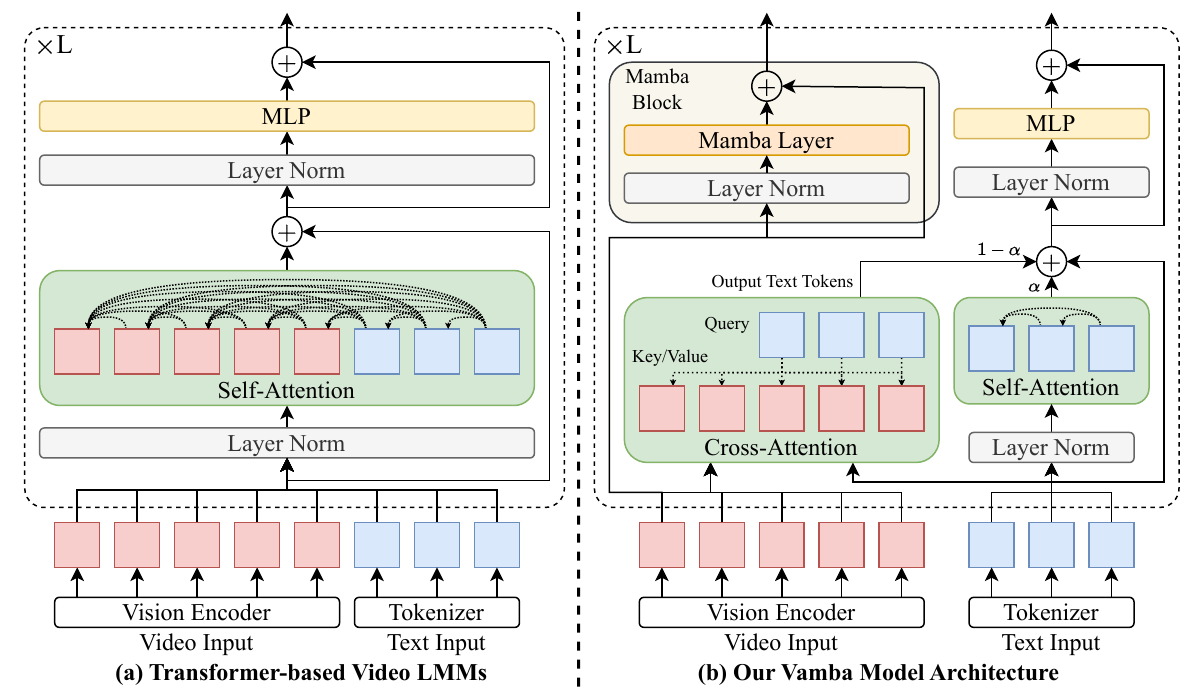}
  \vspace{-1em}
  \caption{Overview of our \model model architecture. Compared to transformer-based LMMs (left), we replace the costly causal self-attention operations with the more efficient cross-attention layers and Mamba blocks to achieve better efficiency.}
  \label{fig:main_attention}
  \vspace{-1em}
\end{figure*}

\subsection{State Space Models and Mamba}
State space models (SSMs) are linear time-invariant systems for modelling continuous signals. A continuous SSM can be expressed by ordinary differential equations (ODEs):
\begin{align*}
    h^\prime(t) = \mathbf{A}h(t) + \mathbf{B}x(t), \quad y(t) = \mathbf{C}h(t),
\end{align*}
here, $x(t)\in \mathbb{R}$ is the input signal, $h(t) \in \mathbb{R}^N$ is the $N$-dimensional hidden state and $y(t) \in \mathbb{R}$ is the output signal. $\mathbf{A}\in \mathbb{R}^{N\times N}$ is the state transition matrix and $\mathbf{B} \in \mathbb{R}^{N \times 1}, \mathbf{C}\in \mathbb{R}^{1\times N}$ are projection matrices.

Mamba \cite{gu2023mamba}, and its predecessor S4 \cite{gu2021efficiently}, are discretized SSMs for discrete sequence modelling. The discretization of the SSM is done by the zero-order hold (ZOH) technique:
\begin{align*}
\begin{split}
    \overline{\mathbf{A}} = \mathrm{exp}(\Delta\mathbf{A}), \quad \overline{\mathbf{B}} = (\Delta \mathbf{A})^{-1}(\mathrm{exp}(\Delta\mathbf{A}) - \mathbf{I})\cdot \Delta \mathbf{B},
\end{split}
\end{align*}
where $\overline{\mathbf{A}},\overline{\mathbf{B}}$ are discretized state variables and $\Delta$ is the step size. The discretized SSM can thus be rewritten as:
\begin{align*}
    h_t = \overline{\mathbf{A}}h_{t-1} + \overline{\mathbf{B}}x_t, \quad y_t = \mathbf{C}h_t.
\end{align*}

Mamba proposes the selective scan algorithm that dynamically determines $\mathbf{B}, \mathbf{C}, \Delta$ based on the sequence inputs. Following S4, the $\mathbf{A}$ matrix is initialized as the diagonalized HiPPO matrix \cite{gu2020hippo}. These design choices are proven to be efficient and effective for modelling extremely long sequences. Mamba-2 \cite{dao2024transformers} further simplifies the formulation of $\mathbf{A}$ to a scalar times identity structure. Mamba-2 supports multi-head SSM and allows a much larger state dimension $N$ than Mamba, while being faster during training.

\subsection{Transformer-based LMMs}
\label{sec:autoregressive_lmm}
We study how text and video tokens are being updated in transformers. As shown in Figure~\ref{fig:main_attention}, we assume a standard case where video tokens are placed before text tokens and ignore the chat template tokens for simplicity. In each LMM decoder layer, the video and text tokens are updated as:
\begin{itemize}
    \item All video tokens get updated by an input layer normalization (LN) layer, a self-attention layer, a post LN layer and a final MLP layer. There are also residual connections for self-attention and MLP layers. The $i^{\mathrm{th}}$ video token $v_i$ computes self-attention based on the first $i$ video tokens:
    \begin{equation}
    \label{equ:causal_self_attention_video}
        o_{v_i}=(\sigma(\frac{q_{v_i} \mathbf{K}_{[v_1:v_i]}^\top}{\sqrt{d}})\mathbf{V}_{[v_1:v_i]})\mathbf{W}_{o},
    \end{equation}
    where $\sigma(\cdot)$ denotes the softmax operation, $q_{v_i}$ is the query vector for $v_i$ and $\mathbf{K}_{[v_1:v_i]}$, $\mathbf{V}_{[v_1:v_i]}$ are key and value matrices of the first $i$ video tokens. $d$ denotes the dimension of the query and key vectors and $\mathbf{W}_{o}$ is the out projection matrix. $o_{v_i}$ represents the final output for token $v_i$.
    \item The text tokens share a similar update route to the video tokens. The self-attention updates are slightly different: for the $j^{\mathrm{th}}$ text token $t_j$, it computes self-attention based on all video tokens and the first $j$ text tokens:
    \begin{equation}
    \label{equ:causal_self_attention_text}
        o_{t_j}=(\sigma(\frac{q_{t_j} [\mathbf{K}_v,\mathbf{K}_{[t_1:t_j]}]^\top}{\sqrt{d}})[\mathbf{V}_v,\mathbf{V}_{[t_1:t_j]}])\mathbf{W}_{o},
    \end{equation}
    where $[\mathbf{K}_v,\mathbf{K}_{[t_1:t_j]}]$ and $[\mathbf{V}_v,\mathbf{V}_{[t_1:t_j]}]$ are the key and value matrices of the combination of all video tokens and text tokens up to $t_j$.
\end{itemize}

The main computation overhead in the transformer-based LMMs comes from the \textbf{quadratic complexity of the self-attention in the video tokens}. To overcome this issue, we design a hybrid Mamba Transformer architecture to process text and video tokens differently. Our detailed model architecture designs are listed in the sections below.

%% file: sec/3_method.tex
\section{Our Method: \model}
Our goal is to devise a model $\Theta^\prime$ that preserves the performance of transformer $\Theta$, while being more efficient. We approach this problem by approximating the causal transformer layers in the pretrained video LMMs.

\subsection{Updating Text Tokens via Cross-Attentions}
\label{sec:cross_attention}
The key idea of our method is to split the expensive self-attention operation over the entire video and text token sequence into two more efficient components. Since video tokens typically dominate the sequence while text tokens remain few, we maintain the self-attention mechanism exclusively for the text tokens and eliminate it for the video tokens. Instead, we add cross-attention layers that use text tokens as queries and video tokens as keys and values. As shown in Figure~\ref{fig:main_attention}, this design updates the text tokens during the attention layers in our model as follows:
\vspace{-0.5em}
\begin{equation}
\label{equ:our_attention}
\begin{aligned}
    o_{t_j}= & (1-\alpha)(\sigma(\frac{q_{t_j} \mathbf{K}_v^\top}{\sqrt{d}})\mathbf{V}_v)\mathbf{W}^{c}_{o} & \text{\small(Cross-Attention)}\\ 
    & +\alpha(\sigma(\frac{q_{t_j} \mathbf{K}_{[t_1:t_j]}^\top}{\sqrt{d}})\mathbf{V}_{[t_1:t_j]}])\mathbf{W}^{s}_{o}, & \text{\small(Self-Attention)}
\end{aligned}
\end{equation}
where $\mathbf{W}_o^c$ and $\mathbf{W}_o^s$ are the output projection matrices for the cross- and self-attention layers. $\alpha\in [0, 1]$ is a learnable weighting parameter that balances the cross- and self-attention outputs. Assuming a total of $M$ video tokens and $N$ text tokens, the self-attention operations in the transformer-based video LMMs have a (pre-filling) computational complexity of $O(d(M+N)^2)$. Our self$+$cross-attention design has the complexity of $O(dN^2)$ (self-attention) and $O(dMN)$ (cross-attention), effectively reducing the pre-filling complexity to $O(dN^2+dMN)\sim O(dMN)$, given that $M\gg N \approx d$.

Our attention formulation can be viewed as an approximation of the causal self-attention operation (Equation~\ref{equ:causal_self_attention_text}), as each text token $q_{t_j}$ can still attend to all video tokens and the first $j$ text tokens. As the video and text tokens share the same channel dimension, the projection matrices in the cross-attention layer will have identical dimensions as in the self-attention layers. This allows us to consider two model design choices: we can either randomly initialize the weights in the cross-attention layers, or we can initialize the cross-attention layers using the self-attention layer weights (i.e. $\mathbf{W}_o^c=\mathbf{W}_o^s$, same applies to the query, key and value projection matrices $\mathbf{W}_q^c,\mathbf{W}_k^c,\mathbf{W}_v^c$). \textit{We add these two design choices to the \model design space.}

\subsection{Updating Video Tokens with Mamba Blocks}
While the self$+$cross-attention design significantly reduces the model's complexity, relying solely on cross-attention layers can compromise the model's representational power. Specifically, rather than updating video tokens through self-attention and MLP layers (i.e., through the causal transformer block, \textit{c.f.} Section~\ref{sec:autoregressive_lmm}), the video tokens remain unchanged after the cross-attention layers. To address this limitation, we seek an efficient architecture to approximate the effects of the transformer blocks. Motivated by the success of the Mamba \cite{gu2023mamba, dao2024transformers} architectures in image and video modelling \cite{zhu2024vision, liu2024vmamba, li2024videomamba}, we propose to employ Mamba blocks to effectively process the video tokens. As shown in Figure~\ref{fig:main_attention}, the video token updates can be formulated as:
\begin{equation}
\label{equ:mamba}
    o_{v_i} = \mathrm{Mamba}(\mathrm{LN}(v_i), \mathbf{h}_{v_{i-1}}, \overline{\mathbf{A}}, \overline{\mathbf{B}}, \mathbf{C}),
\end{equation}
where $\mathrm{LN}(\cdot)$ is the layer normalization operator and $\mathbf{h}_{v_{i-1}}$ is the context (hidden states) at token $v_{i-1}$. As a recurrent model, Mamba operates similarly to causal self-attention layers, enabling each video token to update itself based on all preceding tokens. Crucially, Mamba reduces the complexity of updating video tokens from $O(dM^2)$ in causal self-attentions to $O(d^2M)$. Combining the Mamba blocks and the self$+$cross-attention design, our model achieves an overall complexity of $O(dN^2+dMN+d^2M)\sim O(dMN+d^2M)$, which is substantially lower than the quadratic complexity $O(d(M+N)^2)$ in the transformer-based video LMMs. In Section~\ref{sec:training_paradigm}, we further detail how the Mamba layers are trained to approximate Equation~\ref{equ:causal_self_attention_video}.

\textit{We consider employing Mamba or Mamba-2 blocks in Equation~\ref{equ:mamba} in the \model design space.}

\subsection{Training Paradigm}
\label{sec:training_paradigm}
We utilize a two-stage training strategy to optimize \model, including a pretraining stage and an instruction-tuning stage. In the pretraining stage, the model is initialized with pretrained transformer-based LMM weights in all modules except for the newly introduced cross-attention and Mamba layers. We freeze the pretrained components and train only the new layers using image captioning data to restore the model’s visual understanding capabilities. We explore two types of training objectives during pretraining: (1) the standard language modelling loss $\mathcal{L}_{\mathrm{LM}}=-\frac{1}{T}\sum_{t=1}^{T}\log p(x_t|x_{<t})$, which is the cross-entropy loss for next-token ($x_t$) prediction; (2) the distillation loss $\mathcal{L}_{\mathrm{Distill}}$, where we extract the top 100 logits $\mathcal{P}_{\Theta}$ with the highest values from the transformer-based model $\Theta$ and compute a KL-divergence loss with the logit outputs $\mathcal{P}_{\Theta^\prime}$ of our model $\Theta^\prime$ at the same indices, such that $\mathcal{L}_{\mathrm{Distill}}=D_{KL}(\mathcal{P}_{\Theta}||\mathcal{P}_{\Theta^\prime})$
($D_{KL}(\cdot||\cdot)$ denotes KL-divergence). The final loss is formulated as $\mathcal{L}=\mathcal{L}_{\mathrm{LM}} + \lambda\mathcal{L}_{\mathrm{Distill}}$, where we use $\lambda$ to balance the weights of the two losses. \textit{We include $\lambda=0, 0.001, 0.1, 0.5, 1, 2$ to the \model design space.}

In the instruction-tuning stage, we leverage both image and video instruction-following data to fully finetune \model, thereby enhancing its instruction-following capability. When GPU memory limitations prevent full finetuning, we freeze the vision encoder and finetune only the LMM decoder. We only employ the language modelling loss during the instruct-tuning stage to ensure that the teacher model does not restrict the student's performance.

\textit{The full design space for \model includes whether to initialize cross-attention weights from self-attention; whether to use Mamba or Mamba-2 blocks to update video tokens, and the choice of $\lambda$ in the pretraining stage.}

%% file: sec/4_experiment.tex
\section{Experiments}
In this section, we first ablate the \model design choices and then adopt the best setup for full-scale training.
\subsection{Ablation Study}
We explore the \model design space to determine the effectiveness of our model components. We initialize all our models based on a strong pretrained transformer-based LMM Qwen2-VL-7B \cite{wang2024qwen2}. For all ablation experiments, we randomly select one million images from CC12M \cite{changpinyo2021conceptual} and employ the PixelProse \cite{singla2024pixels} captions for pretraining. We use the LLaVA-Video-178K \cite{zhang2024video} dataset with a total of 1.3M video-instruction pairs for instruction-tuning. Detailed hyperparameter settings can be found in Appendix~\ref{sec:additional_implementation}.

\input{tables/ablation_main}
\input{tables/ablation_distill}

\vspace{-1em}
\paragraph{Evaluation Metrics} For pretraining (stage 1) evaluation, we sample 500 images from the COCO-2017 \cite{chen2015microsoft} dataset and have the models generate captions for each image. We then apply the reference-free version of G-VEval \cite{tong2024g}, a metric based on GPT-4o \cite{OpenAI_GPT4o} to evaluate the quality of the captions. For instruction-tuning (stage 2), we evaluate model performances across three benchmarks: LVBench \cite{wang2024lvbench} for hour-long videos, Video-MME \cite{fu2024video} for medium-to-long videos, and MVBench \cite{li2024mvbench} for short videos. Ablation study results are shown in Table~\ref{tab:ablation_main} and Table~\ref{tab:ablation_distill}.

\vspace{-1em}
\paragraph{Cross-Attention Layer Initialization Strategy} According to Table~\ref{tab:ablation_main}, we find that it is crucial to initialize the cross-attention layer weights using the corresponding self-attention layer from the same decoder level. Compared to Model A, the dramatic performance boost observed in Model B across all metrics underscores the critical impact of this weight initialization strategy. We believe this observation stems from the fact that initializing the cross-attention layer weights from the self-attention layers allows our self$+$cross-attention operation (Equation~\ref{equ:our_attention}) to more closely approximate the pretrained causal self-attention operation (Equation~\ref{equ:causal_self_attention_text}): once $\mathbf{W}_q^c,\mathbf{W}_k^c,\mathbf{W}_v^c$ and $\mathbf{W}_o^c$ are set equal to $\mathbf{W}_q^s,\mathbf{W}_k^s,\mathbf{W}_v^s$ and $\mathbf{W}_o^s$, the only discrepancy between the two equations becomes their attention score matrices. This difference can be mitigated by our two-stage training process, making it easier for Model B to recover its multimodal understanding capabilities.
\input{tables/hour_long}

\vspace{-1em}
\paragraph{Mamba Block Design Choices} Comparing Model B with Model C and D in Table~\ref{tab:ablation_main}, we observe that both Mamba and Mamba-2 blocks improve the model performances across all metrics, indicating that Mamba blocks bring extra representation power to the model's visual understanding capabilities. Furthermore, when comparing Model C and Model D, we find that Mamba-2 demonstrates superior image and video modelling capabilities. Despite having a simplified $\mathbf{A}$ matrix structure, Mamba-2 accommodates a larger state dimension than Mamba (64 versus 16 in our case), which likely contributes to its improved performance.

\vspace{-1em}
\paragraph{Pretraining Distillation Loss} We employ the pretrained Qwen2-VL-7B model as the teacher model and perform a series of pretraining experiments with the additional distillation loss based on our best model setting (Model D in Table~\ref{tab:ablation_main}). The results are shown in Table~\ref{tab:ablation_distill}. In contrast to previous findings from cross-attention-based LLMs (e.g. CEPE \cite{yen2024long}), where distillation loss is reported to be beneficial, we observe that incorporating an additional distillation loss does not further improve our model's performance, as the G-VEval score decreases for all $\lambda > 0$. Therefore, we exclude the distillation loss and only employ the language modelling loss in both stages in our final training setting.

\input{tables/other_benchmarks}

\subsection{Main Evaluation Results}
\label{sec:main_evaluation_results}

\paragraph{Implementation Details} Based on our ablation study (Table~\ref{tab:ablation_main}), we adopt Model D as our final design and initialize \model from Qwen2-VL-7B. We source $\sim$3M image caption data for pretraining and $\sim$7M image and video QA data for instruction-tuning. The full implementation details can be found in Appendix~\ref{sec:additional_implementation}.

\vspace{-1em}
\paragraph{Hour-Long Video Understanding} We evaluate our model’s ability to handle extremely long videos using two public benchmarks: LVBench \cite{wang2024lvbench} and HourVideo \cite{chandrasegaran2025hourvideo} (development set). We further compose a new benchmark called HourEval by selecting all questions related to videos longer than 30 minutes from Video-MME \cite{fu2024video}, MLVU \cite{zhou2024mlvu} development set, and LongVideoBench \cite{wu2025longvideobench} validation set. The average video lengths for LVBench, HourVideo, and HourEval are 68.4, 47.2, and 54.7 minutes, respectively. We compare our model against four efficient video LMMs: LLaVA-Mini \cite{zhang2025llava}, LongLLaVA \cite{wang2024longllava}, Video-XL \cite{shu2024video} and LongVU \cite{shen2024longvu}. We also include the results from Qwen2-VL-7B (our baseline transformer-based LMM) as a reference.

Experimental results are shown in Table~\ref{tab:hour_long}. Our \model consistently outperforms all efficient video LMMs across the three hour-long video benchmarks, highlighting its exceptional ability to understand and reason over hour-scale videos. Notably, our model surpasses the baseline Qwen2-VL-7B on the LVBench benchmark, and its performance on HourVideo is also very close to Qwen2-VL-7B. These results underscore that our \model is competitive with the best open-sourced transformer-based LMMs, while being significantly more efficient during training and inference.

\vspace{-1em}
\paragraph{Medium-Length or Short Video Understanding}
To further demonstrate our model's generalizability across various video durations, we test \model on several medium-length or short video understanding benchmarks. Specifically, we report multiple-choice question performance on Video-MME, MLVU, LongVideoBench, NExT-QA \cite{xiao2021next} and MVBench \cite{li2024mvbench}. We also evaluate \model on DREAM-1K \cite{wang2024tarsier} for video captioning assessment.

According to Table~\ref{tab:other_bench}, our \model demonstrates superior performance across three medium-length video understanding benchmarks (with average video durations between 10–20 minutes), ranking first among efficient video LMMs on all metrics. Despite the newly integrated cross-attention and Mamba layers in \model have only been trained on relatively smaller datasets, \model's performance remains competitive with large-scale pretrained transformer-based LMMs. Furthermore, all our evaluations can be conducted on a single 80G GPU, while some transformer-based LMMs, such as Qwen2-VL-7B, require additional inference strategies like sliding window attention or ring attention over multiple GPUs to achieve optimal results.

For short video understanding and video captioning benchmarks, \model also achieves competitive performances, ranking first on NExT-QA and DREAM-1K and second on MVBench among efficient LMMs. Overall, our model delivers the best results on medium-length and long video benchmarks, demonstrating its strong ability to handle long-context video-language inputs.

\begin{figure}[t!]
    \centering
    
    \begin{subfigure}[b]{1\linewidth}
        \centering
        \caption{Training GPU Memory Usage Comparison}
        \vspace{-2pt}
        \includegraphics[width=\linewidth]{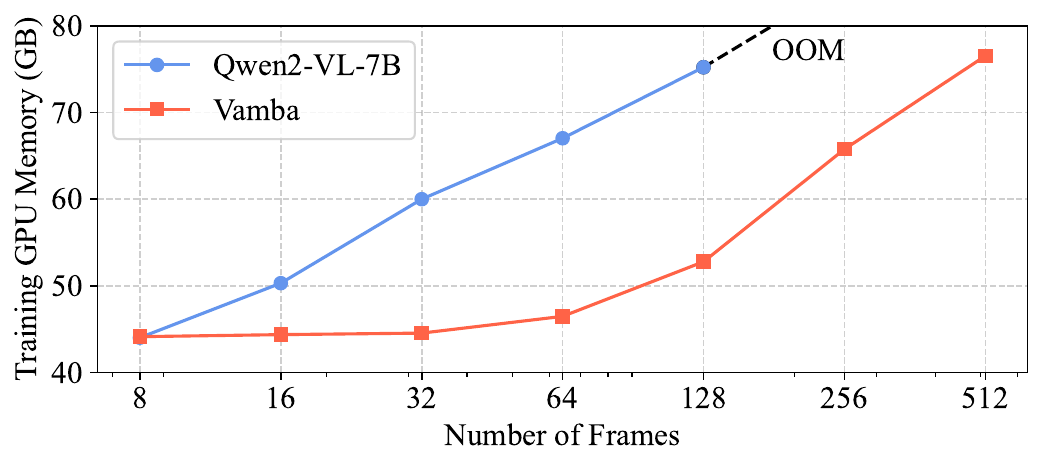}
        \label{fig:train_memory}
    \end{subfigure}
    
    \vspace{-1.5em}
    
    \begin{subfigure}[b]{1\linewidth}
        \centering
        \caption{Training Runtime Comparison (time for 1 training step)}
        \vspace{-2pt}
        \includegraphics[width=\linewidth]{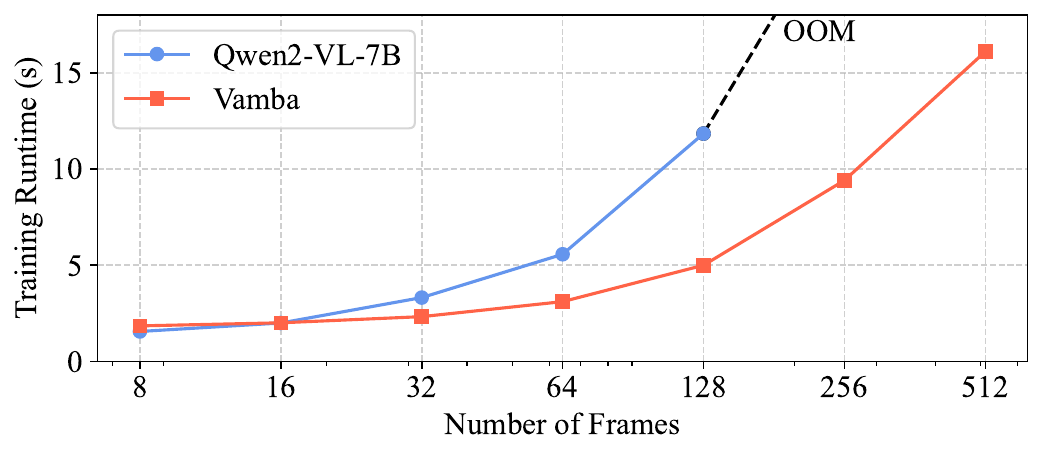}
        \label{fig:train_runtime}
    \end{subfigure}
    \vspace{-3.5em}
    \caption{Comparison of training GPU memory usage and runtime per training step between Qwen2-VL-7B and \model.}
    \label{fig:efficiency_main}
    \vspace{-1em}
\end{figure}

\subsection{Runtime Efficiency Analysis}
To understand our model’s runtime efficiency gains over the baseline transformer-based LMM (Qwen2-VL-7B), we conduct an efficiency analysis for both training and inference. For model training, we ensure a consistent training environment by using 8 NVIDIA A800 80G GPUs for both models. All input videos are resized to a resolution of $640\times360$, and a batch size of 1 is maintained on each GPU. We apply standard optimization methods such as Flash-Attention 2 \cite{dao2023flashattention}, DeepSpeed ZeRO-3 \cite{rajbhandari2020zero} and gradient checkpointing \cite{chen2016training} for both models. Based on different numbers of input frames, we measure the average GPU memory across 8 GPUs and the runtime per training step during training. As shown in Figure~\ref{fig:train_memory}, although \model and Qwen2-VL-7B share the same vision encoder design and generate vision tokens of identical sequence lengths, our model requires over 50\% less training memory when processing videos with more than 16 frames. This efficiency gain allows us to handle a larger number of video frames during training (512 vs. 128). Furthermore, as depicted in Figure~\ref{fig:train_runtime}, our efficient design also accelerates model training, achieving nearly twice the speedup per training step when working with more than 64 frames.

\begin{figure}[t!]
    \centering
    
    \begin{subfigure}[b]{1\linewidth}
        \centering
        \caption{Pre-filling GPU Memory Usage Comparison}
        \vspace{-2pt}
        \includegraphics[width=\linewidth]{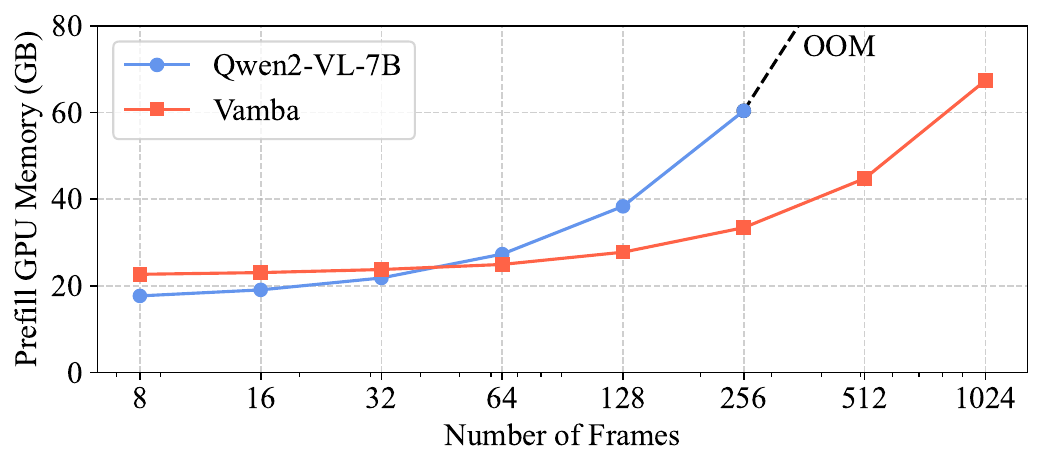}
        \label{fig:memory}
    \end{subfigure}
    
    \vspace{-1.5em}
    
    \begin{subfigure}[b]{1\linewidth}
        \centering
        \caption{Pre-filling FLOPs Comparison (measured using \texttt{calflops})}
        \vspace{-2pt}
        \includegraphics[width=\linewidth]{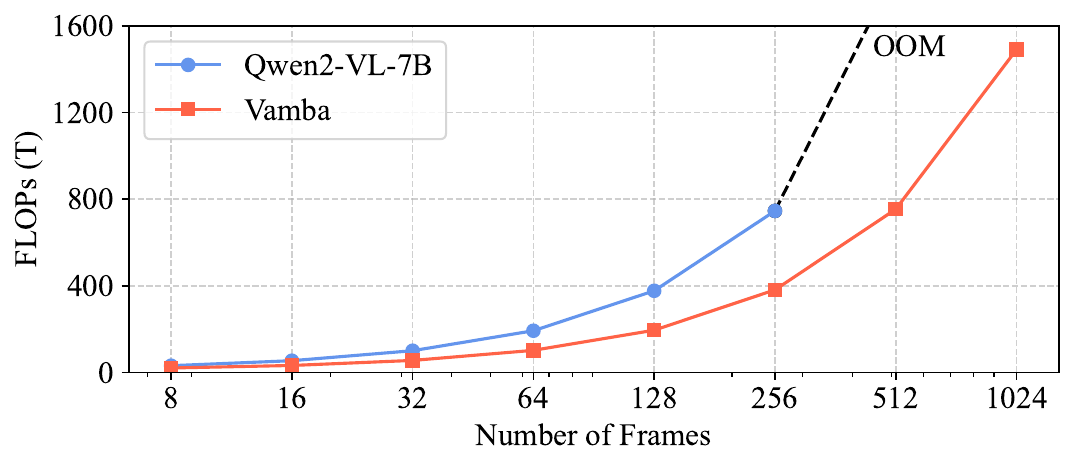}
        \label{fig:flops}
    \end{subfigure}
    \vspace{-3.5em}
    \caption{Comparison of GPU memory usage and FLOPs between Qwen2-VL-7B and \model during inference.}
    \label{fig:efficiency_main}
    \vspace{-1em}
\end{figure}

For model inference, we focus on the pre-fill stage and analyze GPU memory usage and FLOPs for both models. All measurements are conducted on an input video with a resolution of $640\times 360$. As shown in Figure~\ref{fig:memory}, \model requires slightly more GPU memory than Qwen2-VL-7B when the number of input frames is low (fewer than 32 frames), due to the additional $\sim$3B parameters in its cross-attention and Mamba layers. However, \model's memory usage increases more slowly as the frame increases, allowing it to handle four times as many frames on a single NVIDIA A800 80G GPU compared to Qwen2-VL-7B (256 vs. 1024). Regarding computational cost, \model reduces FLOPs by 30\% to 50\% during inference (Figure~\ref{fig:flops}), demonstrating a significantly lower complexity than its transformer-based LMM counterparts.

\begin{figure*}[ht!]
  \centering
  \includegraphics[width=1.0\textwidth]{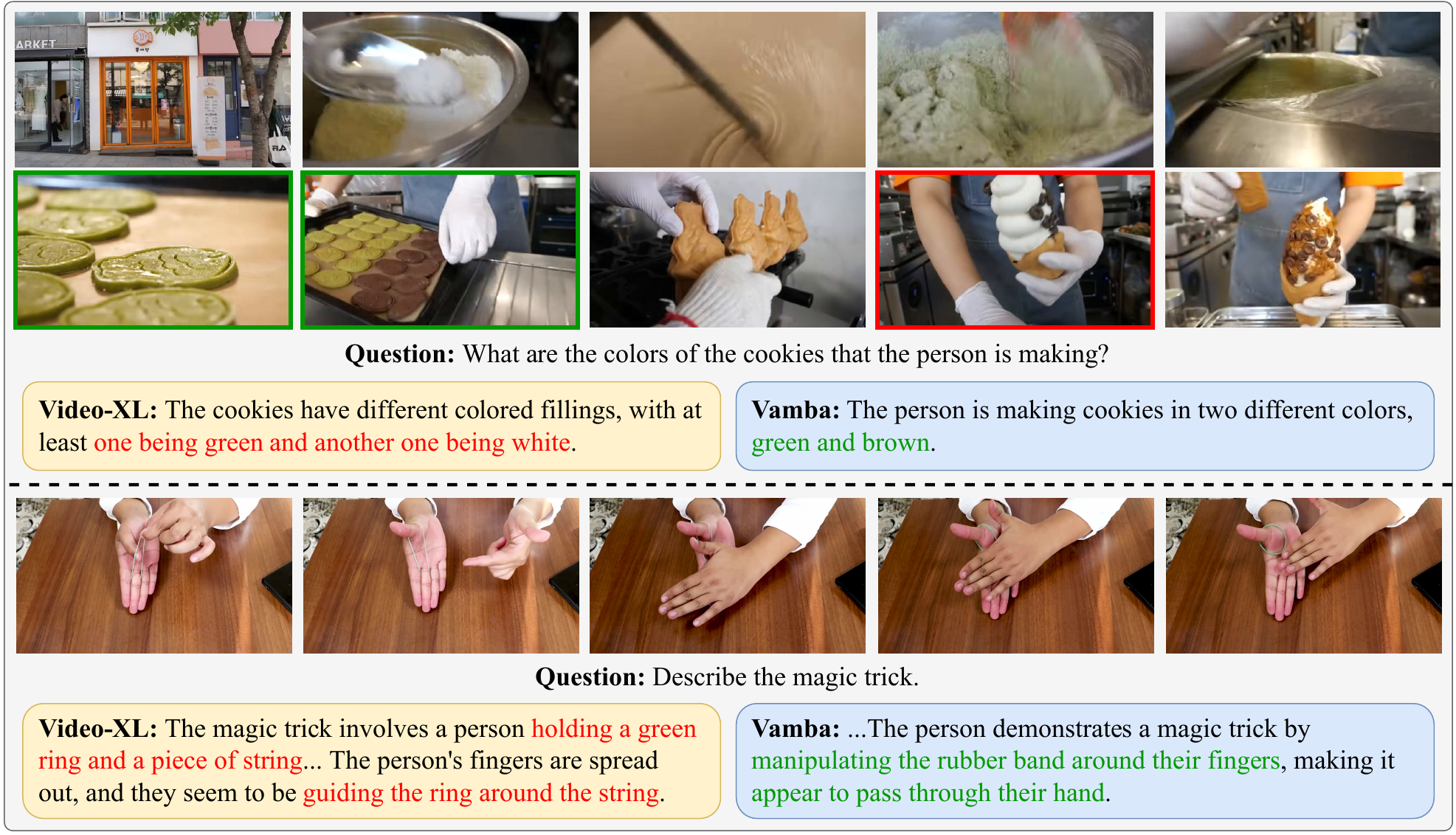}
  \vspace{-2em}
  \caption{Qualitative comparison between \model and efficient LMM baselines. \textcolor{red}{Red text} denotes incorrect responses, while \textcolor{deepgreen}{green text} highlights the correct responses by our \model.}
  \label{fig:case_study}
  \vspace{-1.5em}
\end{figure*}

\subsection{Case Study}
In this section, we perform a case study for Video-XL and \model using different video inputs and show the results in Figure~\ref{fig:case_study}. The ``cookie'' example focuses on object property identification in a long video (36 minutes). Video-XL incorrectly identifies the cookie colors as green and white, possibly due to the video also containing scenes for making other desserts (e.g. white ice cream in the frame highlighted in red). On the other hand, \model more accurately describes them as green and brown. In the ``magic trick'' example, Video-XL misinterprets the rubber band as a green ring, whereas \model correctly describes the performer manipulating the rubber band around their fingers, making it appear to pass through. These comparisons highlight how \model is better at retrieving relevant events in long videos and capturing fine-grained visual details and actions.

%% file: tables/ablation_main.tex
\begin{table}[]
\centering
\small
\caption{Ablation study results for the \model design space explorations. ``CA from SA?'' denotes whether to initialize cross-attention weights from self-attention layers. ``VidMME'' represents the Video-MME benchmark.}
\setlength\tabcolsep{1.5 pt}
\begin{tabular}{@{}ccccccc@{}}
\toprule
\multirow{4}{*}{\begin{tabular}[c]{@{}c@{}}Model\\ ID\end{tabular}} & \multirow{4}{*}{\begin{tabular}[c]{@{}c@{}}CA\\ from\\ SA?\end{tabular}} & \multirow{4}{*}{\begin{tabular}[c]{@{}c@{}}Mamba\\ Block\\ Type\end{tabular}} & Stage 1 & \multicolumn{3}{c}{Stage 2}                                     \\ \cmidrule(l){4-7} 
                                                                    &                                                                          &                                                                               & G-VEval     & \multicolumn{1}{c}{LVBench} & \multicolumn{1}{c}{VidMME}    & MVBench \\ \cmidrule(l){4-7} 
                                                                    &                                                                          &                                                                               & ref. free   & \multicolumn{1}{c}{test}    & \multicolumn{1}{c}{w/o sub.} & test    \\ \midrule
A                                                                   & \xmark                                                                        & N/A                                                                           & 75.7        & \multicolumn{1}{c}{23.7}    & \multicolumn{1}{c}{47.6}         & 40.9    \\
B                                                                   & \cmark                                                                        & N/A                                                                           & 81.0        & \multicolumn{1}{c}{34.2}    & \multicolumn{1}{c}{51.7}         & 51.8    \\
C                                                                   & \cmark                                                                        & Mamba                                                                         & 81.8        & \multicolumn{1}{c}{34.2}        & \multicolumn{1}{c}{53.4}             & 53.5        \\
D                                                                   & \cmark                                                                        & Mamba-2                                                                       & 82.2        & \multicolumn{1}{c}{35.3}    & \multicolumn{1}{c}{54.1}         & 53.5    \\ \bottomrule
\end{tabular}
\label{tab:ablation_main}
\end{table}

%% file: tables/ablation_distill.tex
\begin{table}[]
\centering
\small
\caption{Ablation study results on pretraining \model models with a transformer-based LMM teacher and a distillation loss.}
\setlength\tabcolsep{6.5 pt}
\begin{tabular}{@{}lclcccc@{}}
\toprule
\multirow{2}{*}{\begin{tabular}[c]{@{}l@{}}Model D\\ + Distill\end{tabular}} & \multicolumn{6}{c}{$\lambda$ value in Stage 1 Training}                                                                                                            \\ \cmidrule(l){2-7} 
                                                                             & \multicolumn{1}{c}{0}     & \multicolumn{1}{l}{0.001} & \multicolumn{1}{c}{0.01}  & \multicolumn{1}{c}{0.5}   & \multicolumn{1}{c}{1}     & 2     \\ \midrule
G-VEval                                                                      & \multicolumn{1}{c}{82.19} & \multicolumn{1}{l}{81.05} & \multicolumn{1}{c}{80.68} & \multicolumn{1}{c}{73.69} & \multicolumn{1}{c}{63.65} & 47.61 \\ \bottomrule
\end{tabular}
\label{tab:ablation_distill}
\vspace{-1em}
\end{table}

%% file: tables/hour_long.tex
\begin{table}[]
\centering
\small
\caption{Comparisons between baseline models and \model on hour-long video understanding benchmarks. \textbf{Bold}: best results among efficient video LMMs. \uline{Underline}: second-best.}
\setlength\tabcolsep{3.6 pt}
\begin{tabular}{lcccc}
\toprule
                         &                        & \multicolumn{3}{c}{Hour-Long Video Understanding} \\ 
\cmidrule(l){3-5} 
                         &                        & LVBench                              & HourVideo        & HourEval \\
\cmidrule(l){3-5} 
\multirow{-4}{*}{Models} & \multirow{-4}{*}{Size} & test                                 & dev              & test \\
\midrule
\multicolumn{5}{c}{\cellcolor[HTML]{EFEFEF}\textit{Transformer-based LMMs}} \\
Gemini-1.5-Pro \cite{team2024gemini} & -                      & { 33.1}          & { 37.3}          & {-}          \\
GPT-4o \cite{OpenAI_GPT4o}           & -                      & { 34.7}          & { 19.6}          & {-}          \\
Qwen2-VL \cite{wang2024qwen2}        & 7B                     & { 42.0}          & { 33.8}          & { 53.0}     \\
\midrule
\multicolumn{5}{c}{\cellcolor[HTML]{EFEFEF}\textit{Efficient LMMs}} \\
LLaVA-Mini \cite{zhang2025llava}   & 7B                    & { 17.6}          & { 20.6}          & { 24.2}          \\
LongLLaVA \cite{wang2024longllava} & 9B                    & { 31.2}          & { 27.7}          & { 39.1}          \\
LongVU \cite{shen2024longvu}       & 7B                     & { {\ul 37.8}}    & { 30.8}          & { 46.8}          \\
Video-XL \cite{shu2024video}       & 7B                     & { 36.8}          & { {\ul 33.0}}    & { {\ul 47.1}}    \\
\midrule
\model    & 10B                    & { \textbf{42.1}} & { \textbf{33.6}} & { \textbf{50.7}} \\
\bottomrule
\end{tabular}
\label{tab:hour_long}
\vspace{-1em}
\end{table}

%% file: tables/other_benchmarks.tex
\begin{table*}[!t]
\centering
\small
\caption{Comparisons between baseline models and \model on medium-length or short video understanding benchmarks. \textbf{Bold}: best results among each section. \uline{Underline}: second-best. * indicates the results based on our evaluation scripts.}
\begin{tabular}{lcccccccc}
\toprule
& & \multicolumn{3}{c}{Medium-Length Understanding} & \multicolumn{2}{c}{Short Video Understanding} & Video Captioning \\ \cmidrule(l){3-8} 
Models & Size & Video-MME & MLVU & {LongVideoBench} & MVBench & {NExT-QA} & DREAM-1K \\ \cmidrule(l){3-8} 
& & w/o subtitles & m-avg & {val} & test & {mc} & F1 \\ \midrule
\multicolumn{8}{c}{\cellcolor[HTML]{EFEFEF}\textit{Proprietary Models}} \\
GPT-4V \cite{achiam2023gpt} & - & 59.9 & 49.2 & {59.1} & 43.5 & 70.4 & 34.4 \\
GPT-4o \cite{OpenAI_GPT4o} & - & 71.9 & 64.6 & {66.7} & - & {76.0} & 39.2 \\
Gemini-1.5-Pro \cite{team2024gemini} & - & 75.0 & 62.9 & {64.0} & 54.2 & 76.4 & 36.2 \\ \midrule
\multicolumn{8}{c}{\cellcolor[HTML]{EFEFEF}\textit{Open-source Transformer-based LMMs}} \\
VideoChat2 \cite{li2024mvbench} & 7B & 39.5 & 47.9 & {39.3} & 51.9 & {78.6} & 26.6 \\
ShareGPT4Video \cite{chen2025sharegpt4video} & 7B & 39.9 & 46.4 & {39.7} & 51.2 & - & 19.5 \\
LongVA \cite{zhang2024long} & 7B & 52.4 & 56.3 & {51.8} & 49.2 & {68.3} & - \\
Video-CCAM \cite{fei2024video} & 9B & 50.3 & 58.5 & - & 64.6 & - & - \\
Kangaroo \cite{liu2024kangaroo} & 8B & 56.0 & 61.0 & {54.8} & 61.1 & - & - \\
InternVL2 \cite{chen2024internvl} & 8B & 54.0 & ~48.1\textsuperscript{*} & ~51.8\textsuperscript{*} & 66.4 & - & 26.9 \\
LLaVA-OneVision \cite{li2024llavaov} & 7B & 58.2 & ~62.6\textsuperscript{*} & \textbf{56.4} & 56.7 & \textbf{79.4} & \textbf{31.7} \\
Qwen2-VL \cite{wang2024qwen2} & 7B & \textbf{63.3} & \textbf{~64.2\textsuperscript{*}} & ~52.4\textsuperscript{*} & \textbf{67.0} & - & 29.6 \\
Phi-4-Mini \cite{abouelenin2025phi4minitechnicalreportcompact} & 5.6B & 55.0  & 60.1  & 46.7 & 60.4 & - & -\\
\midrule
\multicolumn{8}{c}{\cellcolor[HTML]{EFEFEF}\textit{Open-source Efficient LMMs}} \\
LLaVA-Mini \cite{zhang2025llava} & 7B & 40.3 & 44.3 & ~19.3\textsuperscript{*} & 44.5 & ~47.6\textsuperscript{*} & 22.9 \\
LongLLaVA \cite{wang2024longllava} & 9B & 51.6 & ~53.3\textsuperscript{*} & ~42.1\textsuperscript{*} & 54.6 & ~72.2\textsuperscript{*} & \uline{24.6} \\
LongVU \cite{shen2024longvu} & 7B & ~55.3\textsuperscript{*} & \uline{65.4} & ~53.5\textsuperscript{*} &  \textbf{66.9} &  \uline{~78.0\textsuperscript{*}} & \textbf{28.1} \\
Video-XL \cite{shu2024video} & 7B & \uline{55.5} & 64.9 & {50.7} & 55.3 & ~77.5\textsuperscript{*} & 23.5 \\
\midrule
\model & 10B & \textbf{57.8} & \textbf{65.9} & {\textbf{55.9}} & \uline{60.4} & \textbf{78.1} & \textbf{28.1} \\ \bottomrule
\end{tabular}
\label{tab:other_bench}
\vspace{-1em}
\end{table*}

%% file: sec/2_related_work.tex
\section{Related Work}
\label{sec:related_work}
\paragraph{Large Multimodal Models (LMMs)}
Large Multimodal Models (LMMs) have rapidly evolved in recent years. Recent research has focused on improving LMM's instruction-following ability through better data curation \cite{llavanext2024, li2024llavaov, chen2024sharegpt4v, li2024mvbench, ren2024vista, zhang2024video}, building better vision encoders that can process images at higher resolutions \cite{llavanext2024, laurenccon2024matters, wang2024qwen2}, and extending LMMs to interleaved image \cite{laurenccon2024matters, li2024llava, jiang2024mantis} or video \cite{maaz2023video, lin2023video, zhang2024long} understanding. To enhance instruction-following capabilities, LLaVA-1.5 \cite{liu2024improved} and LLaVA-NeXT \cite{llavanext2024} leverage GPT-4-generated multimodal data for knowledge-enhanced tuning. To optimize visual feature extraction, Idefics2 \cite{laurenccon2024matters} leverages the NaViT \cite{dehghani2023patch} strategy and the SigLIP \cite{zhai2023sigmoid} vision encoder to support image inputs with native resolution. Qwen2-VL \cite{wang2024qwen2} applied Multimodal RoPE (M-RoPE) to optimize feature extraction for varying input sizes.

Despite significant advances, current state-of-the-art LMMs remain inefficient at handling long-context multimodal inputs due to the quadratic complexity of causal self-attention operations. To mitigate this issue, methods like Flamingo \cite{alayrac2022flamingo}, OpenFlamingo \cite{awadalla2023openflamingo}, Idefics \cite{laurenccon2023obelics} and LLaMA 3.2 \cite{dubey2024llama} interleave gated cross-attention layers with self-attention layers for vision-text modelling. mPlug-Owl3 \cite{ye2024mplug} develops the hyper-attention layer that integrates self-attention with cross-attention to support multimodal understanding. However, these models generally underperform compared to LLaVA-like transformer-based LMMs. In this study, we identify that one possible reason is the lack of vision token updates for cross-attention-based models. Our \model addresses this limitation by incorporating additional Mamba layers to update vision tokens.

\vspace{-1em}
\paragraph{Efficient LLMs/LMMs}
To improve the efficiency of LMM reasoning, recent work \cite{li2023videochat, zhang-etal-2023-video, li2024mvbench, li2023blip, zhang2025llava, shu2024video, shen2024longvu} develops techniques to reduce the length of the vision token sequence. Video-LLaMA \cite{zhang-etal-2023-video} and VideoChat \cite{li2023videochat, li2024mvbench} use a Q-Former \cite{li2023blip} to compress dense visual tokens into a more compact sequence. LLaVA-Mini \cite{zhang2025llava} further reduces each image to only one token. Video-XL \cite{shu2024video} introduces a Visual Summarization Token to summarize the visual information, while LongVU \cite{shen2024longvu} designs a spatiotemporal adaptive compression mechanism to reduce the number of video tokens.

Another research avenue, exemplified by the Mamba \cite{gu2023mamba, dao2024transformers} model series, focuses on developing linear-time sequence models to reduce complexity. Mamba-based \cite{zuo2024falcon} and Mamba-Transformer hybrid models \cite{lieber2024jamba, glorioso2024zamba} have achieved notable success in the realm of pure language modelling. However, these efficient architectures have received little attention in the multimodal understanding domain. LongLLaVA \cite{wang2024longllava} took an initial approach by incorporating the hybrid Jamba \cite{lieber2024jamba} model as the LMM decoder, yet its performance in long video understanding remains suboptimal. This shortfall is likely due to the absence of a strong hybrid LLM as the base model and the lack of large-scale training on multimodal instruction-following data.

%% file: sec/5_conclusion.tex
\section{Conclusion}
We presented \model, a hybrid Mamba-Transformer model for efficient hour-long video understanding. By integrating cross-attention for text tokens and Mamba-2 blocks for video token updates, our approach reduces computational complexity and GPU memory usage while achieving competitive performance across long, medium, and short video benchmarks. Extensive evaluations on datasets such as LVBench demonstrate \model’s superiority over existing efficient video LMMs. For future work, since our model design is orthogonal to token reduction-based methods, we will focus on combining the strengths of both approaches to develop more efficient \model variants.

%% file: sec/X_suppl.tex
\clearpage
\setcounter{page}{1}
\maketitlesupplementary

\section{Additional Implementation Details}
\label{sec:additional_implementation}
We use 8 NVIDIA A800 80G GPUs to train our models for both ablation study and full-scale training. For ablation studies, the learning rate is set to 1e-5 for pretraining and 1e-7 for instruction-tuning. We further conduct a hyperparameter search and find that setting the learning rate to 5e-6 during instruction-tuning works the best for \model across multiple benchmarks. We therefore set the learning rate to 1e-5 for pretraining and 5e-6 for instruction tuning for full-scale training. We employ a cosine learning rate schedule for all training stages in both ablation studies and full-scale training. The training batch size is set to 128. We employ training optimization methods such as Flash-Attention 2 \cite{dao2023flashattention}, DeepSpeed ZeRO-3 \cite{rajbhandari2020zero} and gradient checkpointing \cite{chen2016training} to reduce the training cost, and apply sequence parallelism to pack multiple samples into one sequence during training in both stages.

\section{Model Evaluation Details}
In this section, we provide more details for benchmarking \model and our selected baseline models.

\subsection{Baseline Models}

\textbf{Qwen2-VL} \cite{wang2024qwen2} is an LMM that uses the Qwen2 LLM as its backbone and a DFN-derived Vision Transformer with 2D RoPE positional embedding. It is pretrained on a vast 1.4T-token multimodal corpus composed of image-text pairs, OCR text (images of text), interleaved image–text web articles, visual QA data, video dialogues, and image-based knowledge datasets. The pre-training is staged: 600B tokens for vision-language alignment followed by 800B tokens mixing richer image–text content and VQA/multitask data, alongside continued pure text to maintain language skills. Finally, Qwen2-VL is instruction-tuned via ChatML-format dialogs that span multiple modalities, e.g. document parsing, comparisons of two images, long video understanding, and even agent-oriented visual tasks.

\noindent\textbf{LLaVA-Mini} \cite{zhang2025llava} is a compact multimodal model built on a 7–8B Vicuna LLM with a CLIP ViT-based vision encoder. It uses the same training data as LLaVA-1.5 \cite{liu2024improved}: about 558K image–caption pairs for initial vision–language pre-training and 665K image-grounded instruction examples for fine-tuning. The pre-training stage aligns visual features to text using caption datasets like COCO \cite{chen2015microsoft} and VisualGenome-based \cite{krishna2017visual} captions, while the instruction-tuning stage uses multimodal dialogues. An enhanced variant of LLaVA-Mini further incorporates 100K video-based instruction samples from Video-ChatGPT \cite{maaz2023video} and other open sources, combined with the original 665K image instructions (total ~3M training instances) to extend its capability to video understanding.

\noindent\textbf{LongLLaVA} \cite{wang2024longllava} extends LLaVA \cite{liu2024improved} to handle very long visual contexts by using a hybrid Transformer–Mamba architecture with a Jamba-9B backbone for language. It follows a three-stage training process, including a single-image feature alignment on Allava-Caption \cite{chen2024allava} and ShareGPT4V \cite{chen2024sharegpt4v}, a single-image instruction fine-tuning on LLava-1.5 and Mantis-Single \cite{jiang2024mantis}, and multi-image instruction fine-tuning on VideoChat2 \cite{li2023videochat} and ShareGPT4Video \cite{chen2025sharegpt4video}. By progressively increasing the number of images per sample, LongLLaVA learns temporal and spatial dependencies and can efficiently handle input sequences up to around 1000 images. 

\noindent\textbf{LongVU} \cite{shen2024longvu} is a multimodal model geared toward long video understanding. It first learns from 3.2 million image–text pairs via a single-image training stage using the LLaVA-OneVision dataset \cite{li2024llavaov}. It then leverages a subset of VideoChat2-IT \cite{li2023videochat} that contains around 0.55M videos, 1K video-classification clips from Kinetics-710 \cite{carreira2019short}, and about 85K multimodal video instruction dialogues from the ShareGPT4Video \cite{chen2025sharegpt4video}.  Additionally, the MovieChat long-video dialogue data \cite{song2024moviechat} is used to provide hour-length conversational examples. This rich training mix enables LongVU to handle extended videos by adaptively compressing frames while preserving essential visual details.

\noindent\textbf{Video-XL} \cite{shu2024video} employs an LLaMA-based 7B language model and a CLIP ViT-L vision encoder, and it is trained entirely on image-based data despite targeting long videos. Its two-stage training first performs projection-layer pre-training on 2M image–text pairs from Laion-2M \cite{schuhmann2021laion} to align visual features with the text space. It then undergoes visual instruction tuning on roughly 695K image-grounded instruction samples from Bunny-695k \cite{he2024efficient}, where the model learns to follow image-based instructions. The training approach lets Video-XL handle hour-long videos in context by compressing visual tokens, achieving strong results on benchmarks for long video comprehension.

\subsection{Evaluation Benchmarks}

\noindent\textbf{LVBench} \cite{wang2024lvbench} is a benchmark designed to test the ability of video LMMs to comprehend extremely long videos. It contains 1,549 question-answer pairs, with an average video length of 4,101 seconds. The evaluation focuses on six fundamental aspects: temporal grounding, which involves identifying specific moments in a video; video summarization, which assesses the model's ability to condense key information; video reasoning, which tests logical inference from video content; entity recognition, which identifies people, objects, or places; event understanding, which captures the sequence and significance of events; and key information retrieval, which ensures the model extracts crucial details. The full test set is used for evaluation.

\noindent\textbf{HourVideo} \cite{chandrasegaran2025hourvideo} is a benchmark dataset for long-form video-language understanding, focusing on videos up to one hour in length. It consists of 500 carefully selected first-person videos sourced from the Ego4D \cite{grauman2022ego4d} dataset, with each video ranging from 20 to 120 minutes in duration. The dataset includes 12,976 human-annotated multiple-choice questions covering four major task categories: summarization, perception, visual reasoning, and navigation. HourVideo is designed to challenge models in long-context reasoning and multimodal comprehension across extended video timelines. Benchmark results reveal that existing multimodal models, such as GPT-4 and LLaVA-NeXT, perform only marginally better than random chance, while human experts achieve an accuracy of 85.0\%. This highlights the dataset's difficulty and the current gap in long-video understanding capabilities.

\noindent\textbf{Video-MME} \cite{fu2024video} is a benchmark specifically designed to evaluate how well LMMs can analyze video content. It features a dataset of 900 videos and 2700 questions, covering six different visual domains. The questions are grouped based on video length into short, medium, and long categories, with median durations of 26 seconds, 164.7 seconds, and 890.7 seconds, respectively. The benchmark supports two evaluation methods: (1) the “w/ subtitle” setting, where both subtitles and questions are provided as text inputs, and (2) the “w/o subtitle” setting, which relies only on raw video inputs alongside the questions. Our study primarily focuses on the “w/o subtitle” format to enhance long video comprehension by leveraging video-based augmentation rather than textual cues from subtitles.

\noindent\textbf{MLVU} \cite{zhou2024mlvu} is a benchmark designed to assess long video understanding across various tasks and video genres. It includes two types of questions: multiple-choice and freeform generation. The evaluation framework measures LMM performance in three key aspects: (1) holistic video understanding, which requires comprehending the entire video for global context; (2) single-detail video understanding, which focuses on recognizing key moments or short segments; and (3) multi-detail video understanding, which involves drawing connections between multiple short clips within the video. Our paper specifically reports accuracy scores for multiple-choice questions from the MLVU development set.

\noindent\textbf{LongVideoBench} \cite{wu2025longvideobench} is a question-answering benchmark designed for interleaved long video-text input. It includes 3,763 videos and 6,678 human-annotated multiple-choice questions covering 17 fine-grained categories. The benchmark supports two evaluation formats: (1) the standard format, where video tokens are processed first, followed by the question descriptions, and (2) the interleaved video-text format, where subtitles are inserted between video frames. We evaluate all baseline models and our \model using the standard input format. The reported results are based on the validation split.

\noindent\textbf{NExT-QA} \cite{xiao2021next} is a video question-answering benchmark designed to evaluate reasoning-based video understanding. It consists of 5,440 videos and approximately 52,000 human-annotated question-answer pairs, covering a diverse range of real-world activities. The dataset includes two types of question formats: multiple-choice questions and open-ended free-form questions. NExT-QA emphasizes causal and temporal reasoning, requiring models to understand event sequences, cause-effect relationships, and interactions within videos. The dataset is divided into three categories: causal, temporal, and descriptive questions. The dataset is split into training (3,870 videos), validation (570 videos), and test (1,000 videos), ensuring standardized benchmarking.

\noindent\textbf{MVBench} \cite{li2024mvbench} is a comprehensive multimodal video understanding benchmark. The dataset introduces a novel static-to-dynamic task transformation, converting existing static image tasks into video-based challenges, assessing a model's ability to perform both low-level perception and high-level cognitive reasoning over time. MVBench automatically converts annotations from 11 publicly available video datasets into unified multiple-choice question-answer pairs, covering diverse scenarios ranging from first-person to third-person perspectives and indoor to outdoor environments. Each question presents five answer choices, ensuring standardized evaluation through human-verified ground truth responses.

\noindent\textbf{DREAM-1K} \cite{wang2024tarsier} is a video description dataset designed for fine-grained event and motion understanding. It contains 1,000 short videos, each averaging 9 seconds, and covers a diverse set of real-world and cinematic scenarios. Unlike question-answering datasets, DREAM-1K requires models to generate detailed multi-sentence descriptions that capture all key actions, interactions, and temporal sequences within each video. The dataset includes videos from five different sources—live-action movies, animated films, stock footage, long YouTube videos, and short-form social media clips—ensuring broad coverage of visual styles. DREAM-1K prioritizes event-based reasoning, expecting models to understand sequential actions, motion cues, and interactions rather than just static descriptions. Evaluation is conducted using AutoDQ (Automatic Description Quality), which measures how well generated descriptions align with reference descriptions by comparing extracted events.

\begin{figure*}[!t]
  \centering
  \includegraphics[width=1\textwidth]{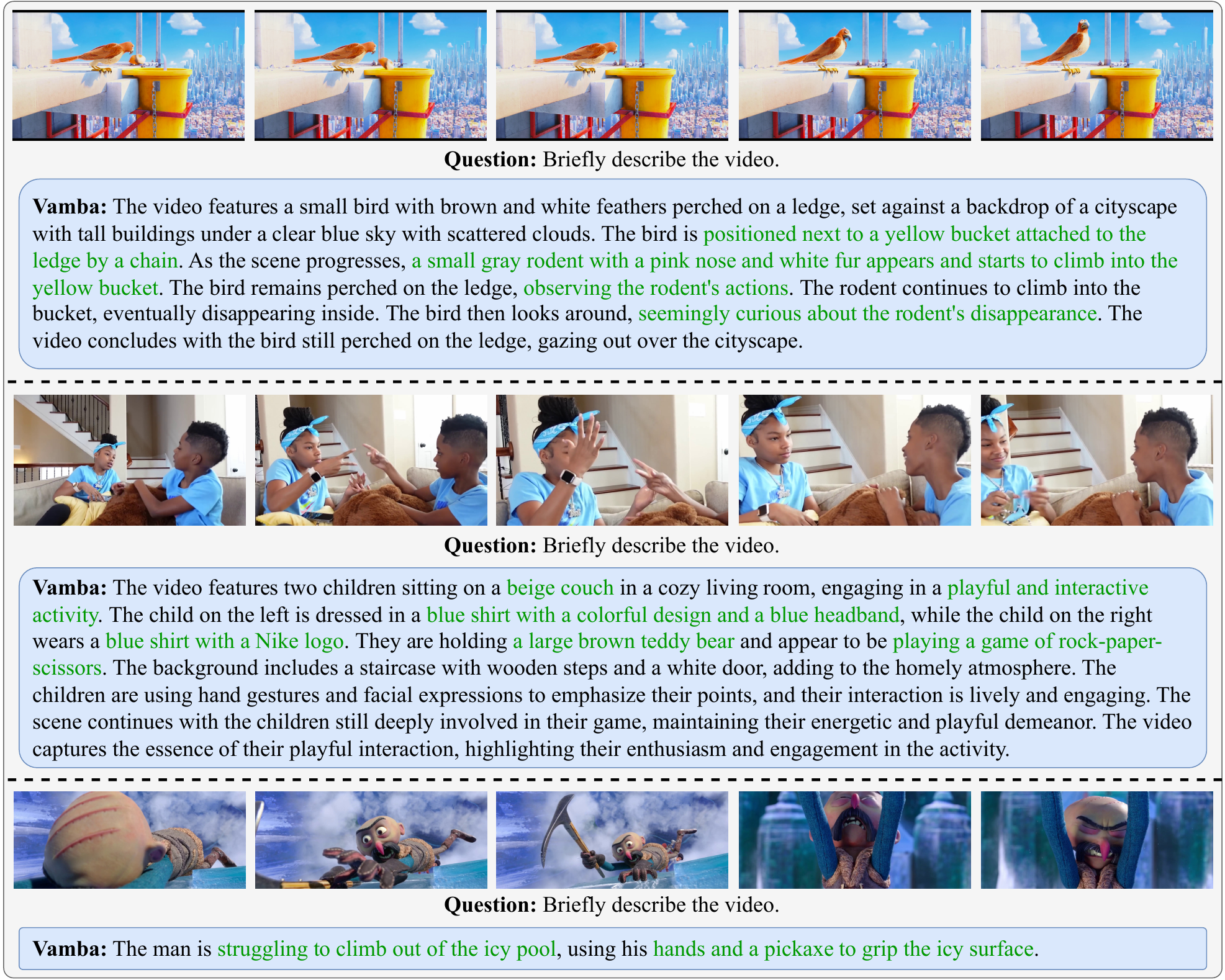}
  \caption{Additional qualitative results for \model.}
  \label{fig:case_study_2}
\end{figure*}

\section{Comparison with Contemporary Work}
Several contemporary works also investigate hybrid Mamba-Transformer models for long video understanding. For example, STORM \cite{jiang2025token} and BIMBA \cite{islam2025bimba} utilize Mamba blocks between the vision encoder and LMM decoder as additional processing and compression modules for video tokens, achieving high performance in long video understanding. However, different from \model, the overall architecture of the LMM remains unchanged in these methods, with the decoder still relying on full causal self-attention layers for both text and video tokens. As a result, the model architectures proposed in these methods offer limited gains in training and inference efficiency, with any speedup in video processing still primarily attributed to token reduction. In comparison, \model directly employs Mamba-2 layers in the LMM decoder and bypasses the self-attention updates for video tokens. This design enables highly efficient video processing even without reducing the number of tokens.

\begin{table}[h!]
\centering
\small
\caption{Quantitative results for \model with token reduction.}
\setlength\tabcolsep{3 pt}
\begin{tabular}{@{}l|c|ccc@{}}
\toprule
Models    & GPU Mem (MB)                  & LVBench                     & VideoMME                     & MLVU                     \\ \midrule
\model    & 45791                         & 42.4                        & 57.4                         & 65.9                     \\
\model-TR & 33847                         & 41.6                        & 56.9                         & 66.5                     \\ \bottomrule
\end{tabular}
\vspace{-10pt}

\label{tab:token_reduction}
\end{table}

\section{Combining \model and Token Reduction}
As mentioned in the paper, we expect \model to be compatible with token reduction, and combining \model and token reduction can potentially result in similar performance and even higher efficiency. We provide some preliminary results for combining \model and token reduction in this section. As shown in Table~\ref{tab:token_reduction}, we can simply uniformly drop 50\% of the video tokens during inference (denoted as \model-TR) and achieve little performance drop across multiple benchmarks with better efficiency. We believe finetuning \model with token reduction can further preserve its capacity and leave this as a future work.

\section{Additional Qualitative Results}
In this section, we showcase more qualitative results from our \model for detailed video captioning and video event understanding. The results are shown in Figure~\ref{fig:case_study_2}.